\documentclass[letterpaper]{article} % DO NOT CHANGE THIS
\usepackage{aaai25}  % DO NOT CHANGE THIS
\usepackage{times}  % DO NOT CHANGE THIS
\usepackage{helvet}  % DO NOT CHANGE THIS
\usepackage{courier}  % DO NOT CHANGE THIS
\usepackage[hyphens]{url}  % DO NOT CHANGE THIS
\usepackage{graphicx} % DO NOT CHANGE THIS
\usepackage{natbib}  % DO NOT CHANGE THIS AND DO NOT ADD ANY OPTIONS TO IT
\usepackage{caption} % DO NOT CHANGE THIS AND DO NOT ADD ANY OPTIONS TO IT
\usepackage{algorithm}
\usepackage{algorithmic}

\usepackage{newfloat}
\usepackage{listings}
\DeclareCaptionStyle{ruled}{labelfont=normalfont,labelsep=colon,strut=off} % DO NOT CHANGE THIS
\floatstyle{ruled}
\newfloat{listing}{tb}{lst}{}
\floatname{listing}{Listing}
\usepackage{amsmath}
\usepackage{amssymb}
\usepackage{enumitem}
\usepackage{booktabs}
\usepackage{multirow}
\title{Integrating Sequence and Image Modeling in Irregular Medical Time Series Through Self-Supervised Learning}
\author{
    Liuqing Chen\textsuperscript{\rm 1,2}, Shuhong Xiao\textsuperscript{\rm 1}, Shixian Ding\textsuperscript{\rm 1}, Shanhai Hu\textsuperscript{\rm 1}, Lingyun Sun\textsuperscript{\rm 1,2}
    \thanks{Corresponding Author}
}
\affiliations{
    \textsuperscript{\rm 1}College of Computer Science and Technology, Zhejiang University, China\\
    \textsuperscript{\rm 2}International Design Institute, Zhejiang University, China \\
    sunly@zju.edu.cn
}

\begin{document}

\maketitle

\begin{abstract}
Medical time series are often irregular and face significant missingness, posing challenges for data analysis and clinical decision-making. Existing methods typically adopt a single modeling perspective, either treating series data as sequences or transforming them into image representations for further classification. In this paper, we propose a joint learning framework that incorporates both sequence and image representations. We also design three self-supervised learning strategies to facilitate the fusion of sequence and image representations, capturing a more generalizable joint representation. The results indicate that our approach outperforms seven other state-of-the-art models in three representative real-world clinical datasets. We further validate our approach by simulating two major types of real-world missingness through leave-sensors-out and leave-samples-out techniques. The results demonstrate that our approach is more robust and significantly surpasses other baselines in terms of classification performance. 
\end{abstract}

\begin{links}
    \link{Code}{https://github.com/zju-d3/AAAI25-Irregular-Medical-Time-Series}
    % \link{Datasets}{https://aaai.org/example/datasets}
    % \link{Extended version}{https://aaai.org/example/extended-version}
\end{links}

\section{Introduction}

Multivariate time series are utilized in various real-world applications, particularly in the medical field, where they are used to record vital signs and laboratory test results for diagnosis \cite{chaudhary2020utilization,brizzi2022spatial}. Typically, these time series are irregular, faced with asynchronicity across sensors and nonuniform sampling in the time domain \cite{chowdhury2023primenet,huang2024dna}. Moreover, significant missing values are usually present in clinical data collection. For example, random missingness can result from patients joining or leaving treatments midway, or complete absence of data from a sensor when specific tests are not conducted \cite{de2019deep}. Some public clinical datasets, such as PhysioNet2012, take even a 80\% missing rate, posing challenges for data analysis and clinical decision-making \cite{wang2024deep}. 

Deep learning methods have been widely adopted to model irregular time series. Some methods rely on the assumption of time discretization, utilizing LSTMs \cite{neil2016phased,weerakody2023policy}, RNNs \cite{che2018recurrent,ma2020adversarial,miao2021generative}, and Transformers \cite{horn2020set,huang2024dna} to capture characteristics of discrete sequences. Nonetheless, these methods often face difficulties in accumulating errors from missing observations \cite{ma2019learning}. Recently, vision models have also shown promising potential in handling irregular sequence data \cite{li2024time}. By transforming series into corresponding RGB representations, visual frameworks can effectively capture dynamic trends and inter-sensor relationships within images \cite{ maroor2024image,li2024time}. However, such designs perform poorly with sparse series that exhibit heavy missing rate \cite{li2024time}. 

We recognize that no one has yet integrated both sequence and image representations in handling irregular medical time series. This introduces a pivotal question: \textit{How can we effectively merge these two distinct representations to improve the robustness of classification for irregular medical time series with extensive missing values?} 

\begin{figure*}[htp]
    \centering
    \includegraphics[width=0.91\linewidth]{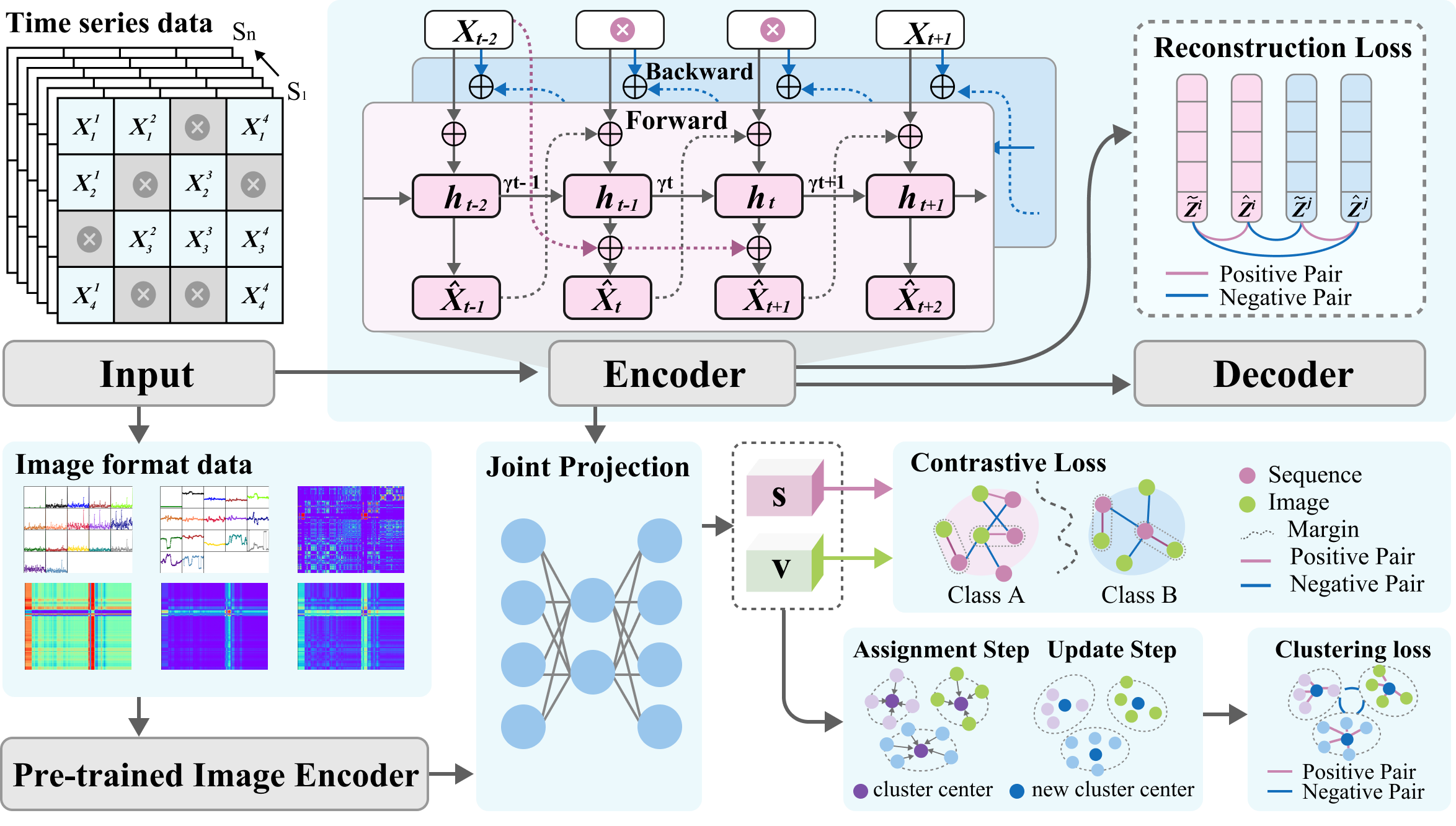}
    \caption{The framework of our approach.}
    \label{framework}
\end{figure*}

To investigate this question, we utilize a joint learning framework that incorporates both sequence and image representations. Additionally, we propose different self-supervised learning (SSL) strategies to enhance the integration and capture of supplementary information across these two representations. Specifically, our approach consists of three main components, as shown in Figure \ref{framework}. For the sequence modeling branch, we employ a generator-discriminator structure and adopt an adversarial strategy \cite{ma2019learning,miao2021generative} for sequence imputation task to minimize the propagation of cumulative errors. In the image branch, we implement different image transformation strategies to improve the performance on sparse series, and utilize a pre-trained Swin Transformer \cite{liu2021swinv2,li2024time} to obtain the corresponding image representations. Three different SSL losses are designed: (1) an inter-sequence contrastive loss to stabilize the sequence imputation process; (2) a sequence-image contrastive loss with margin to learn a more generalizable joint representation for downstream classification; and (3) a clustering loss on joint representations to push similar cases closer across different batches.

We conduct experiments on three real-world clinical datasets: PAM \cite{reiss2012introducing}, P12 \cite{goldberger2000physiobank}, and P19 \cite{reyna2020early}. Table \ref{dataset_statistics} presents their statistics, which show that all three datasets experience severe missing values. We compare our approach with seven other state-of-the-art (SOTA) methods in terms of classification performance. Specifically, our approach achieves the best performance across all three datasets. For the PAM dataset, we observe improvements of 3.1\% in Accuracy, 2.9\% in Precision, 2.3\% in Recall, and 2.6\% in F1 score compared to the second-best method. For the P12 and P19 datasets, we use AUPRC and AUROC as evaluation metrics. Our approach surpasses prior SOTA by 1.1\% (AUPRC) and 0.9\% (AUROC) on P12, and 5.8\% (AUPRC) and 2.3\% (AUROC) on P19. Furthermore, we test further missingness through leave-samples-out and leave-sensors-out experiments on the PAM dataset. In the most severe scenario, with an additional 50\% missing values, our approach demonstrates better robustness, outperforming the second-best method by 6.1\% in Accuracy, 5.9\% in Precision, 3.4\% in Recall, and 4.6\% in F1 score.

The contributions of this paper are summarized as follows:
\begin{itemize} 
\item We propose a joint representation learning framework for multivariate irregular medical time series. To the best of our knowledge, this is the first approach to incorporate both sequence and image modeling.
\item We outline three SSL strategies: inter-sequence contrastive loss, sequence-image contrastive loss, and clustering-based loss. These strategies together enable better integration of sequence and image representations, enhancing the robustness against heavy missingness.
\item Our approach outperforms seven other SOTA methods on three real-world clinical datasets. We also simulates two classic types of missingness and experiments show that our method offers better robustness in handling these cases.
\end{itemize}

\section{Related Work}
\subsection{Irregular Time Series Methods}

Early practices for modeling irregular time series with missing values typically relied on fixed-time discretization. In this context, \cite{choi2016doctor} ignores the timestamp information by treating all intervals as equal, \cite{lipton2016modeling} considers missing data as an effective feature for learning, 
% \cite{futoma2017learning} employs Gaussian processes to model missing data and high uncertainty in real-world situations, 
and \cite{harutyunyan2019multitask} segments the data into evenly spaced time intervals. In contrast, GRU-D \cite{che2018recurrent} employs a gated network and incorporates imputation of missing values into the optimization process. Unlike previous methods, it adopts an additional missing value mask and lag matrix as inputs. Similar strategy have been adopted in \cite{,ma2019learning,ma2020adversarial,miao2021generative}, where adversarial frameworks are utilized to enhance the prediction of imputed values. 

Some recent approaches have leveraged attention mechanisms to improve modeling. For instance, SeFT \cite{horn2020set} introduces a set of differentiable set functions and uses attention mechanisms to aggregate embeddings of different variables. ContiFormer \cite{chen2024contiformer}, on the other hand, combines neural ordinary differential equations (ODEs) with attention mechanisms based on continuous-time dynamics, extending the relationship modeling capabilities of Transformers to the continuous time domain. Besides, DNA-T \cite{huang2024dna} utilizes a deformable attention mechanism to dynamically adjust the receptive field, enabling more effective handling of local features and short-term correlations. Warpformer \cite{zhang2023warpformer} also considers multi-scale features by applying a warping module to achieve multi-grained representations. Unlike previous methods that adopt a sequence modeling perspective, ViTST \cite{li2024time} transforms the signals into RGB images and utilizes a pre-trained Swin Transformer for further classification and regression.

% There are still methods that fall outside the aforementioned categories. One work worth reviewing is Raindrop \cite{zhang2021graph}, which models times series from the perspective of graph neural networks. In this approach, each observation resembles a raindrop hitting a sensor graph, spreading information through a ripple effect. 

% On the other hand, diffusion models \cite{han2022card} are also a growing trend and have been explored in various fields such as energy \cite{xu2024denoising}, finance \cite{daiya2024diffstock}, and microbiology \cite{seki2023imputing}.

\subsection{Modeling Time Series as Images}
Transforming time series data into images has gained significant attention with the advancements in visual detection frameworks. Some approaches \cite{sood2021visual,sangha2022automated,ao2023image, semenoglou2023image, maroor2024image} plot time series directly as time-observation representations and utilize convolutional neural networks (CNNs) for downstream tasks. Generally, they do not apply special processing to the sequences, instead focusing on leveraging visual frameworks to better capture temporal patterns in visualized sequences. ViTST \cite{li2024time} is another similar case that extends further to multivariate sequences and discusses the impact of visualization parameters such as color, markers, and order. 

In contrast, other methods emphasize the modeling of time series, which requires more specialized design and expert knowledge. \cite{tripathy2018use} utilizes an iterative filtering (IF) approach to produce different intrinsic mode functions (IMFs) from EEG signals. Empirically, these transformed features often fit the task better than the original signals. Chong et al. \cite{chong2011signal} and Deng et al. \cite{bs2023_1730} model sequences based on time segmentation, calculating time-invariant features and transforming them into corresponding RGB images. Similarly, frequency domain modeling, as demonstrated by TimesNet \cite{wu2023timesnet}, has also proven effective. By utilizing fast Fourier transform (FFT) to concatenate signal of different time periods, it constructs a 2D representation optimized for CNNs. Finally, other methods model the relative relationships between points in a time series. Examples include Gramian Angular Field (GAF), Markov Transition Field (MTF), and recurrence plot \cite{10.5555/2832747.2832798,hatami2018classification}. Typically, these methods involve applying a reversible time coordinate transformation and calculating the correlations between points, effectively capturing the continuity and periodic characteristics of the sequences.

\section{Approach}

\subsection{Notations}

For a given clinical time series dataset \( D \), each sample \( X \in \mathbb{R}^{d \times T} \) represents a set of \( d \) records over a time \( T = \{t_1,...,t_n\} \), corresponding to a label \( y\). A binary mask \( M \in \mathbb{R}^{d \times T} \) is used to indicate the presence of missing observations in \( X \), where \( M_i^j = 0 \) signifies that the observation of the \( i^{th} \) item at time \( j \) is missing. 

To better handle consecutive missing values time, we follow \cite{miao2021generative,che2018recurrent} to obtain a time-lag matrix \( \delta \in \mathbb{R}^{d \times T} \) for each sample \( X \). This matrix quantifies the time elapsed since the most recent non-missing value for each observation, defined as follows.
\[
\delta^j_i = 
\begin{cases} 
0, &  i = 1 \\
t_i - t_{i-1}, &  m^j_{i-1} = 1 \text{ and } i > 1 \\
\delta^j_{i-1} + t_i - t_{i-1}, &  m^j_{i-1} = 0 \text{ and } i > 1
\end{cases}
\]

For each sample \( X \), the corresponding image \( I \) is constructed, where \( I \in \mathbb{R}^{3\times W \times H} \) represent a certain RGB format image. In total, we implement six transformed images as shown in Figure \ref{framework}. The specific transformation methods applied are as follows: Line Graphs, Frequency Spectrums, Gramian Angular Summation/Difference Fields, Markov Transition Fields, Recurrence Plots.

\subsection{The Model Overview}
In this section, we introduce the overall framework of our model, which comprises three main parts: (a) the sequence encoder, (b) the image encoder, and (c) the joint representation module. The sequence encoder consists of a generator-discriminator pair employing an adversarial strategy for imputation. The generator, \( G \), takes the time series \( X \), the mask \( M \), and the lag matrix \( \delta \) as inputs. Its objective is to estimate the missing values in \( X \) and generate a completed sequence \( X' \). This completed sequence \( X' \) is then used to obtain the sequence representation \( s \in \mathbb{R}^{d} \).  The discriminator D evaluates these estimations with the goal of distinguishing true observations from the imputed values. It outputs a binary matrix \( M' \), which identifies the regions of imputation predicted. For the image encoder, it takes a transformed image \(I \) as input and output the corresponding image representation \( v \in \mathbb{R}^{d}\). Finally, the joint representation module is responsible for mapping the sequence representation \( s \) and the image representation \( v \) into the same space. It then uses the final joint feature \( u \in \mathbb{R}^{d} \) for classification. 

\subsection{Sequence Branch with Imputation}

We adopt a modified bidirectional recurrent neural network (BiRNN) as our generator \(G\), which has been widely used in imputation tasks \cite{ che2018recurrent, ma2019learning, ma2020adversarial, miao2021generative, xu2024learning}. Taking the forward update step as an example, we update the current hidden state as:
\begin{align}
h_t &= \tanh\left(W_h (\gamma_t \odot h_{t-1}) + W_h'(\hat{x}_t + x_\delta )+ b_h\right) \\
\gamma_t &= \exp\left\{-\max(0, W_{\gamma} \delta_t + b_{\gamma})\right\}
\end{align}
In this setup, \(\gamma_t\) is derived from the lag matrix to model the dynamics of decay, where a longer duration of missing data leads \(\gamma_t\) closer to 0. It is applied to determine the extent to which the previous hidden state \(h_{t-1}\) should be retained. In the updating process of \( h_t \), instead of solely utilizing the previous reconstruction \(\hat{x}_t\) as done in prior works, we introduce an additional computation involving \( x_{\delta} \) as Eq. \ref{decay}. 
\begin{align}
\label{decay}
x_\delta = x_{t^-} \cdot \exp \left\{ -\max(0, W_\delta \delta_t) + b_\delta \right\}
\end{align}
This assumes the closest observation \( x_{t^-} \) prior to the current missing value influences the reconstruction process, with this influence decreasing as the time gap increases. 

Then, using a fully connected layer, the new reconstruction of the next step is obtained as: \(\hat{x}_{t+1} = W_{\hat{x}} h_t + b_{\hat{x}}\). And the overall imputed sequence \( X' \) is represented as: \(X' = M \odot X + (1 - M) \odot avg(\hat{X}_{for} + \hat{X}_{back})\), where we take the average of forward and backward result, and only the missing parts are replaced. Finally, the sequence representation \( s \) is obtained as: 
\begin{align}
s = Drop(W_s \cdot LayerNorm(X') + b_s)
\end{align}
In particular, \(W_h\), \(W_h'\), \(W_\gamma\), \(W_\delta\), \(W_{\hat{x}}\), \(W_s\), \(b_h\), \(b_x\), \(b_\gamma\), \(b_\delta\), \(b_{\hat{x}}\), and \(b_s\) are learnable parameters of the model and \( \odot \) denotes the element-wise multiplication. 

We formulate the objective of generator \(G\) into two components: adversarial loss and reconstruction loss. The adversarial loss is defined as the standard GAN's \cite{goodfellow2020generative}:
\begin{align}
\mathcal{L}_{adv} = \mathbb{E} [ (1 - M) \log (1 - D(X'))]
\end{align}
For the reconstruction loss, previous methods often use regression-based metrics such as mean square error (MSE) \cite{ma2020adversarial} or mean absolute error (MAE) \cite{ma2019learning} to assess the consistency between the missing and imputed sequences. However, when dealing with severely missing data, these strategies often fail to model the underlying data patterns, force the generator to learn nothing during the adversarial training phase. Inspired by \cite{raghu2023sequential}, we adopt a self-learning strategy to construct our reconstruction loss, and one choice is the normalized temperature-scaled cross-entropy loss (NT-Xent) \cite{chen2020big}. Given 2\(B\) pairs \((z_i, z_j)\) totally, it is computed as: 
\begin{align}
\label{NT}
\mathcal{L}_{NT} = \frac{1}{2B} \sum_{i=1}^{2B} -\log \frac{\exp(sim(z_i, z_j) / \tau)}{\sum_{k=1}^{2B} 1_{[k \neq i]} \exp(sim(z_i, z_k) / \tau)}
\end{align}
where cosine similarity is used as \(sim(z_i, z_j)\) and \(\tau\) is the temperature hyperparameter. We use NT-Xent to enforce consistency between the forward and backward predictions, as well as between the original and imputed sequences. Thus, the reconstruction loss is defined as: 
\begin{align}
\mathcal{L}_{rec} =  \mathcal{L}_{NT}(\hat{X}_{for}, \hat{X}_{back})\ + \mathcal{L}_{NT}(X, X')
\end{align}

We employ the same RNN in \cite{ma2019learning} as our discriminator \(D\), which takes \(X'\) as input and determines whether each observation is generated with a binary matrix \(M'\). Therefore, the discriminator is trained by minimizing:
\begin{align}
\label{dis}
\mathcal{L}_{dis} = \mathbb{E}[Mlog M' + (1-M)log(1-M')]
\end{align}

\subsection{Imaging Time Series}
We use a pre-trained Swin Transformer \cite{liu2021swinv2} as our image encoder. For the given image input \(I\), the Swin Transformer constructs a hierarchical representation to integrate both local and global information. Specifically, at earlier layers, it partitions the input into small patches and progressively merges neighboring patches as depth increases. It employs two types of attention mechanisms: window-based multi-head self-attention (W-MSA) and shifted window multi-head self-attention (SW-MSA). These mechanisms are respectively used to compute self-attention within a fixed window and to calculate dynamic relationships between windows. The vectors from the last stage after layer normalization are used as our image representation \( v \in \mathbb{R}^{d}\).

Overall, we implement six types of images for representation learning and a detailed description is presented in Appendix A. 

\begin{itemize}[leftmargin=*]
\item \textbf{Line Graphs} are constructed as \cite{li2024time}, with each variable represented by a line image of uniform size. 
\item \textbf{Frequency Spectrums} are generated based on the Fourier transform, considering that frequency domain signals tend to be more robust in cases of extreme data missingness.
\item \textbf{Gramian Angular Fields} \cite{10.5555/2832747.2832798} transform time series into polar coordinates, constructing trigonometric sums/ differences between any two time points to represent temporal correlation. 
\item \textbf{Markov Transition Fields} record the Markov transition probabilities between any two time observations \cite{10.5555/2832747.2832798}. They are insensitive to the distribution of the time series and temporal step information, allowing them to effectively capture correlations between observations with substantial missing data.
\item \textbf{Recurrence Plots} \cite{hatami2018classification}, based on phase space reconstruction, transform time series data into trajectories within phase space and analyze their recurrences. They are designed to  capture the inherent repetitiveness and periodicity within the time series.
\end{itemize}

\subsection{Joint Representations Through Contrast and Clustering}

The joint representation module includes a transformation function \( f: s,v \to \mathbb{R}^D
 \), which projects and concatenates the sequence features \( s \) and image features \( v \) into a joint space \( R^D \), and the fused feature is obtained as \( u = [s,v] \). To ensure both the quality and consistency of the joint representation, we implement contrastive learning within each batch to maximize the mutual information between corresponding pairs. A simple choice is to use the NT-Xent in Eq. \ref{NT}, where only sequence and image features corresponding to the same sample are treated as positive pairs \cite{sangha2024biometric}. Through this approach, NT-Xent ensures that the similarity between representations from the same sample is higher than that of other pairs. However, it also misses opportunities to learn from a wider set of potential pairs \cite{li2022clustering}. 
 
 In this case, a step forward is to treat \(s\) and \(v\) from different samples within the different category as a special form of negative pairs, thereby enhancing the model's ability to distinguish inter-class differences. Specifically, we introduce an additional margin \(m\) for these special negative pairs, enforce the model to exert greater effort to distinguish them:

% \noindent\resizebox{\columnwidth}{!}{
% \begin{minipage}{0.6\textwidth} 
% \begin{align}
% -\frac{1}{B} \sum_{i=1}^{B} &\left[ \log \left(\frac{\exp((v_i \cdot s_i)/ \tau)}{\sum_{j \in P(i)} \exp((v_i \cdot s_j)/ \tau) + \sum_{j \notin P(i)} \exp((v_i \cdot s_j + m)/ \tau)}\right)\right. \nonumber \\
% & + \log \left(\frac{\exp((v_i \cdot s_i)/ \tau)}{\sum_{k \in P(i)} \exp((v_k \cdot s_i)/ \tau) + \sum_{k \notin P(i)} \exp((v_k \cdot s_i + m)/ \tau)}\right]
% \end{align}
% \label{cont}
% \end{minipage}
% }

\begin{equation}
\scalebox{0.93}{$
-\frac{1}{B} \sum_{i=1}^{B} \left[ \log \left( \frac{\exp((v_i \cdot s_i)/ \tau)}{\sum_{j \in P(i)} \exp((v_i \cdot s_j)/ \tau) + \sum_{j \notin P(i)} \exp((v_i \cdot s_j + m)/ \tau)} \right) \right. \nonumber$}
\end{equation}
\begin{equation}
\scalebox{0.93}{$
\left. + \log \left( \frac{\exp((v_i \cdot s_i)/ \tau)}{\sum_{k \in P(i)} \exp((v_k \cdot s_i)/ \tau) + \sum_{k \notin P(i)} \exp((v_k \cdot s_i + m)/ \tau)} \right] 
\right.$}
\label{cont}
\end{equation}

Here, for the \(i^{th}\)sample, \(P(i)\) represents the set of all sample index that are in the same category. 

In contrastive learning, the formation of positive and negative pairs is confined to each batch. However, this approach lacks control over the semantic relationships between samples across different batches. As a result, similar samples from separate batches may not receive similar representations. In this case, we incorporate clustering learning into the training process to push semantically similar samples together across batches.

Specifically, we applied the K-means algorithm to the fused feature \(u\). We begin with the assignment step: during each training epoch, we select a set of k (k $\ll$ N) representative features \([C_{u_1}, \ldots, C_{u_k}]\) as the cluster centers for that round. Each fused feature \(u_i\) is assigned to a set \(S_k\) with center \(C_{u_k}\) by minimizing the overall distance as defined in Eq. \ref{cluster}.

\begin{align}
\label{cluster}
\underset{S}{\mathrm{arg min}}  \sum_{j=1}^{k} \sum_{u_i \in S_j} \|u_i - C_{u_j}\|^2 
\end{align}

% \begin{align}
% \mathcal{L}_{\text{cluster}} = -\frac{1}{N} \sum_{i=1}^{N} \min_{j=1 \ldots k} \left( 1 - \cos(X_i, C_j) \right),
% \end{align}

We then use these cluster centers, \([C_{u_1}, \ldots, C_{u_k}]\), as contrastive loss reference targets to construct the clustering loss:
\begin{align}
\mathcal{L}_{cluster} = -\frac{1}{N} \sum_{i=1}^{N} \log \frac{\exp(\cos(u_i, C_{u_i})/\tau)}{\sum_{j=1}^{k} \exp(\cos(u_i, C_{u_j})/\tau)}
\end{align}
where a cluster center \(C_{u_k}\) and all elements within  set \(S_k\) are treated as positive pairs, and elements from different clusters are considered negative pairs. To ensure sufficient samples for optimizing clustering, we perform the update step at the end of each epoch: we iteratively update the cluster centers using Eq. \ref{update} and calculate new assignments with Eq. \ref{cluster}, until the total distance is less than a predefined threshold \(\tau_c\).

\begin{align}
\label{update}
C_k = \underset{u \in S_k}{\mathrm{arg min}} \sum_{u' \in S_k} \|u - u'\|^2 
\end{align}

\subsection{Overall Training Process}

The overall training process is divided into three steps as follows:

\begin{itemize}[leftmargin=*]
    \item Firstly, we fix the generator \( G \) and update the discriminator \( D \) based on Eq. \ref{dis} .
    \item Next, we update the parameters of \( G \) based on the new \( D \), with the objective function \(  \mathcal{L}_{adv} +  \alpha \mathcal{L}_{rec} \).
    \item Finally, we compute the forward pass of all three components, utilizing the joint feature \( u \) to perform classification. For the PAM dataset, we use the Cross Entropy Loss as the classification loss \(\mathcal{L}_{clf}\). For the more imbalanced P12 and P19 datasets, we opt for the Focal Loss. The final objective is expressed as \(\mathcal{L}_{clf} + \beta_1 \mathcal{L}_{cont} + \beta_2 \mathcal{L}_{cluster}\).
\end{itemize}

% \begin{algorithm}[H]
% \caption{Training Process}
% \label{alg:training}
% \begin{algorithmic}[1]
% \STATE \textbf{Initialize:} Generator $G$, Discriminator $D$, and all parameters.

% \STATE \textbf{Step 1: Update Discriminator}
% \STATE Fix $G$
% \FOR{each training step}
%     \STATE Update $D$ based on Eq. (8)
% \ENDFOR

% \STATE \textbf{Step 2: Update Generator}
% \STATE Fix $D$
% \FOR{each training step}
%     \STATE Update $G$ with the objective: $L_{adv} + \alpha L_{rec}$
% \ENDFOR

% \STATE \textbf{Step 3: Joint Forward Pass and Classification}
% \FOR{each forward pass}
%     \STATE Compute joint feature $u$
%     \STATE Perform classification using $u$
%     \IF{dataset is PAM}
%         \STATE Compute classification loss $L_{clf}$ using Cross Entropy Loss
%     \ELSIF{dataset is P12 or P19}
%         \STATE Compute classification loss $L_{clf}$ using Focal Loss
%     \ENDIF
% \ENDFOR

% \STATE \textbf{Final Objective:}
% \STATE $L_{final} = L_{clf} + \beta_1 L_{cont} + \beta_2 L_{cluster}$
% \end{algorithmic}
% \end{algorithm}

\section{Experiments}

\subsection{Datasets and Metrics}

\begin{table}[h]
\centering
% \resizebox{0.98\columnwidth}{!}{ 
\setlength{\tabcolsep}{0.8mm}
\begin{tabular}{@{}lccccr@{}} 
\toprule
\textbf{Dataset} & \textbf{Features} & \textbf{Time} & \textbf{Classes} & \textbf{Missing Ratio} & \textbf{Samples} \\ 
\midrule
PAM              & 17                & 600           & 8                & 60\%                   & 5,333             \\
P12              & 36                & 215           & 2                & 88.4\%                 & 11,988            \\
P19              & 34                & 60            & 2                & 94.9\%                 & 38,803            \\
\bottomrule
\end{tabular}
% }
\caption{Statistics of datasets utilized.}
\label{dataset_statistics}
\end{table}

\begin{table*}[htp]
\setlength{\heavyrulewidth}{1.3pt}
\centering
\begin{tabular}{c|cccc|cc|cc}
\toprule
\multirow{2}{*}{Methods}  & \multicolumn{4}{c|}{PAM} & \multicolumn{2}{c|}{P12} & \multicolumn{2}{c}{P19}  \\ 
\cmidrule{2-9} 
 & Accuracy & Precision & Recall & F1 score & AUROC & AUPRC  & AUROC & AUPRC  \\
\midrule
GRU-D & 83.3 {$\pm$ \scriptsize 1.6}  & 84.6 {$\pm$ \scriptsize 1.2} & 85.2 {$\pm$ \scriptsize 1.6}  & 84.8 {$\pm$ \scriptsize 1.2} &  81.7 {$\pm$ \scriptsize 1.8} & 41.3 {$\pm$ \scriptsize 3.5}  & 83.6 {$\pm$ \scriptsize 2.1}  &  45.7 {$\pm$ \scriptsize 4.2}   \\
SeFT  & 63.3 {$\pm$ \scriptsize 2.2} & 66.7 {$\pm$ \scriptsize 2.4}  & 65.3 {$\pm$ \scriptsize 1.5} & 65.1 {$\pm$ \scriptsize 1.8} & 73.3 {$\pm$ \scriptsize 2.5} & 29.1 {$\pm$ \scriptsize 4.1} & 84.5 {$\pm$ \scriptsize 2.3} & 46.7 {$\pm$ \scriptsize 3.1}   \\
% SSGAN       &  &  &  &  &  &  &  &    \\
CARD      &71.9 {$\pm$ \scriptsize 2.9}  &75.5 {$\pm$ \scriptsize 2.8}  & 73.5 {$\pm$ \scriptsize 3.1} & 73.8 {$\pm$ \scriptsize 3.0} &71.4 {$\pm$ \scriptsize 0.9}  &26.1 {$\pm$ \scriptsize 1.2}  &80.7 {$\pm$ \scriptsize 1.0}  & 36.7 {$\pm$ \scriptsize 6.0}    \\
Raindrop       & 89.2 {$\pm$ \scriptsize 1.3} & 90.8 {$\pm$ \scriptsize 1.0}  &90.4 {$\pm$ \scriptsize 1.3}  &90.5 {$\pm$ \scriptsize 1.2} & 82.0 {$\pm$ \scriptsize 2.4} & 44.3 {$\pm$ \scriptsize 3.3}  & 82.7 {$\pm$ \scriptsize 3.9} &52.3 {$\pm$ \scriptsize 3.9}   \\
PrimeNet      &85.5 {$\pm$ \scriptsize 1.5}  & 87.8 {$\pm$ \scriptsize 1.2}  & 87.1 {$\pm$ \scriptsize 1.1} & 87.1 {$\pm$ \scriptsize 1.2} &\underline{85.1} {$\pm$ \scriptsize 0.8}  &\underline{49.3} {$\pm$ \scriptsize 1.9} &80.3 {$\pm$ \scriptsize 0.5}  &31.6 {$\pm$ \scriptsize 0.9}    \\
ContiFormer &66.6 {$\pm$ \scriptsize 1.8}   &68.6 {$\pm$ \scriptsize 1.7}   &69.7 {$\pm$ \scriptsize 1.5}   & 67.4 {$\pm$ \scriptsize 1.7}  & 72.1 {$\pm$ \scriptsize 0.4}  & 29.6 {$\pm$ \scriptsize 0.8}  & 80.7 {$\pm$ \scriptsize 0.3}  & 34.7 {$\pm$ \scriptsize 1.9}   \\
ViTST       &\underline{95.2} {$\pm$ \scriptsize 1.4}  &\underline{95.8} {$\pm$ \scriptsize 1.3}  & \underline{96.1} {$\pm$ \scriptsize 1.1} & \underline{95.9} {$\pm$ \scriptsize 1.2} & 84.2 {$\pm$ \scriptsize 1.1} & 43.2 {$\pm$ \scriptsize 2.4} &\underline{89.3} {$\pm$ \scriptsize 0.2}  &\underline{53.8} {$\pm$ \scriptsize 1.1}    \\
\midrule
Ours       &\textbf{98.3} {$\pm$ \scriptsize 0.3}  &\textbf{98.7} {$\pm$ \scriptsize 0.6}  & \textbf{98.4} {$\pm$ \scriptsize 1.0} & \textbf{98.5} {$\pm$ \scriptsize 0.7} &\textbf{86.0} {$\pm$ \scriptsize 0.3}  &\textbf{50.4} {$\pm$ \scriptsize 2.1} &\textbf{91.6} {$\pm$ \scriptsize 0.9} & \textbf{59.6} {$\pm$ \scriptsize 1.3}    \\
\bottomrule
\end{tabular}
\caption{Comparison with state-of-the-art baselines on irregularly sampled time series classification. We use \textbf{bold} to indicate the best results and \underline{underline} for the second best one.}
\label{baselines}
\end{table*}

In the experiments, we consider three real-world irregular clinical datasets as shown in Table \ref{dataset_statistics}. The physical activity monitoring (PAM) dataset \cite{reiss2012introducing} focuses on tracking human activities, containing data from eight person who performed nine different actions. This dataset comprises 5,333 samples and captures data from four types of sensors placed at three distinct body locations, encompassing a total of 17 observational variables. The P12 dataset \cite{goldberger2000physiobank} includes 11,988 patient samples from ICU stays, with 36 measurements each. The binary labels indicate the prognosis for each sample as either survival or not. Finally, the P19 dataset \cite{reyna2020early} contains data from 38,803 sepsis patients, each with 34 measurements, and a high missing rate of 94.9\%. Approximately 90\% of these patients died due to sepsis.

To maintain consistency across all experiments, we follow the same data partition as \cite{zhang2021graph, li2024time}, dividing the datasets into training, validation, and testing sets in an 8:1:1 ratio. For the PAM dataset, we use Accuracy, Precision, Recall, and F1 score as evaluation metrics. For the more imbalanced P12 and P19 datasets, we report the Area Under the ROC Curve (AUROC) and the Area Under the Precision-Recall Curve (AUPRC). For more experimental results that are not included in this section, we present them in Appendix D.

\subsection{Implementation and Training}
We use Gated Recurrent Units (GRU) \cite{dey2017gate} in both our generator and discriminator. The generator has 4 layers, with the number of units fixed at 128. The discriminator is a 5-layer RNN and the number of units is set to $\{$128, 64, 16, 64, 128$\}$, respectively. A checkpoint pre-trained on ImageNet-21K dataset are utilized for our image encoder. The patch size and window size are 4 and 7. For the P12 and P19 datasets, all images are set to a size of 384 $\times$ 384 pixels. While for the PAM dataset, line graph and frequency spectrum are configured to 256$ \times$ 320, while all other images are set to 320 $\times$ 320. We use a 3-layer MLP as our joint projection, with the number of units set to $\{$1024, 512, 1024$\}$. 
% For hyperparameters, the temperature parameter \( \tau \) for both the reconstruction loss, contrastive loss, and clustering loss is set at 1.2. The margin \( m \), \( \alpha \), \( \beta_1 \) and \( \beta_2 \) is specified at 0.05, 4, 0.1, and 0.2, respectively. We discuss the selection of these hyperparameters in Appendix B. 

For the P12 and P19 datasets, the total epoch is set to 8 and we apply upsampling of the minority class to mitigate imbalance. For the PAM dataset, we set the total epoch to 40. The batch sizes used for training are 32 for P19 and P12, and 48 for PAM. For each dataset, we discuss the learning rate as well as more hyperparameter settings in Appendix B. All experiments are performed on a server with NVIDIA GeForce RTX 3090 24GB and PyTorch 2.4.0+cu124. 

\subsection{Results}

% SSGAN \cite{miao2021generative}
\subsubsection{Comparison with state-of-the-art methods.} We compare our approach against seven state-of-the-art methods for irregularly sampled time series, including GRU-D \cite{che2018recurrent}, SeFT \cite{horn2020set}, CARD \cite{han2022card}, Raindrop \cite{zhang2021graph}, PrimeNet \cite{chowdhury2023primenet}, ContiFormer \cite{chen2024contiformer}, and ViTST \cite{li2024time}. For each baseline, we introduce our implementation and hyperparameter settings in Appendix C. To ensure a fair evaluation, we average the performance of each method across five individual tests, using the same data splits and settings provided in \cite{li2024time}. 

Table \ref{baselines} presents the comparison results, highlighting that our approach outperforms the other seven state-of-the-art methods across all three datasets. Specifically, we achieve a significant improvement on the PAM datasets, with an increase of 3.1\% in Accuracy, 2.9\% in Precision, 2.3\% in recall, and 2.6\% in F1 score. For the P12 and P19 datasets, our approach shows improved performance in predicting minority classes, with an increase of 0.9\%, 2.3\% in absolute AUROC points, and 1.1\%, 5.8\%  in absolute AUPRC, respectively. 

% \textcolor{red}{We compare the number of parameters between the baseline methods and ours, as shown in Table 3. It can be seen that the largest model, Raindrop, has a parameter count of 150M and all methods belong to small or medium-sized models. Therefore, it will not result in significant resource consumption. Methods like GRU-D and Contiformer, which only perform sequence modeling, have relatively fewer parameters, whereas methods involving image representation, such as ViTST and ours, have much larger parameter counts.}

\subsubsection{Performance under increased missing rates.}

\begin{figure*}[htp]
    \centering
    \includegraphics[width=0.95\linewidth]{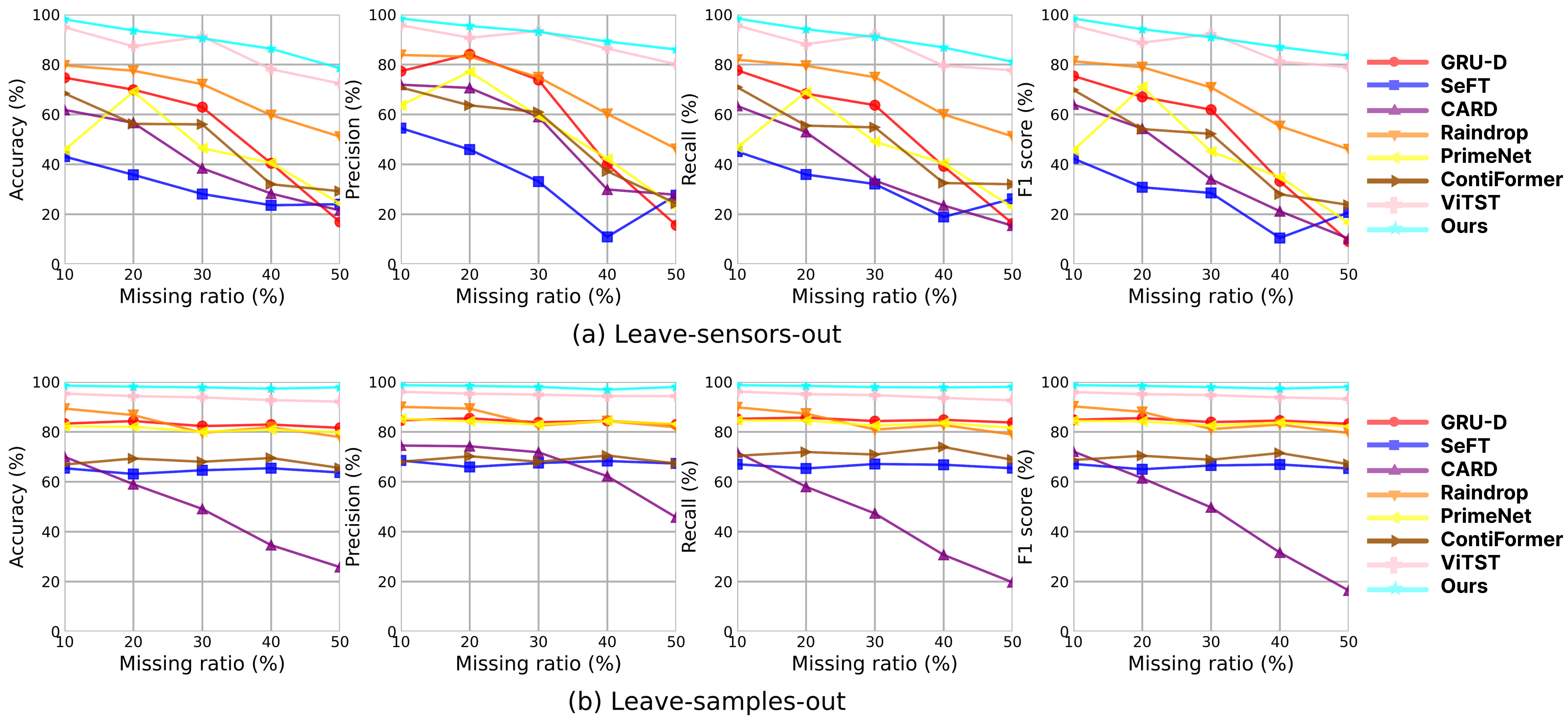}
    \caption{Performance under increased missingness: (a) leave-sensors-out and (b) leave-samples-out on the PAM dataset. Tests are conducted with 10\%-50\% extra missing values.} 
    \label{missing test}
\end{figure*}

To further validate the robustness of our approach, we conduct additional experiments to compare the performance under increased levels of missing rate. Given that the P12 and P19 datasets have already faced very high missing rates—88.4\% and 94.9\% respectively, we conduct all the tests on the PAM dataset, which originally has a missing rate of 60\%. We conducted two types of tests: the leave-sensors-out setting, simulating scenarios where certain medical tests are not performed, and the leave-samples-out setting, reflecting situations where patients join or leave treatments midway. We follow the approach in \cite{zhang2021graph}, applying all modifications only to the test set by randomly masking the original observations.

As shown in Figure \ref{missing test}, our approach consistently achieves the best performance in all settings. For the leave-sensors-out tests, as the missing ratio increase from 10\% to 50\%, our approach exhibit the least performance decline. Even in the most extreme scenario, where 50\% of the sensors (9 sensors) are masked, all our metrics remain above 80\%. Compared to the second-best method, ViTST, our approach outperform by 6.1\%, 5.9\%, 3.4\%, and 4.6\% in Accuracy, Precision, Recall, and F1 score, respectively. The margins are even more significant compared to the third-ranked Raindrop, with improvements of 27.4\%, 39.8\%, 29.9\%, and 37.5\% in the same metrics. For the leave-samples-out setting, we randomly sampled and masked time steps. Overall, only CARD experienced significant decline as missing rate increases, while most models shows relatively minor decline, indicating that they effectively capture the temporal relationships between time steps. In terms of absolute performance, our model outperform the second-best, ViTST, by 5.7\% in Accuracy, 3.8\% in Precision, 5.4\% in Recall, and 4.8\% in F1 score at a 50\% missing rate.

\subsubsection{Clinical Turing tests.}
To ensure that the learned representations align with clinically meaningful patterns rather than statistical artifacts, we conducted a clinical Turing test on the generated signals, as described in \cite{gillette2023medalcare}. Specifically, we select 60 samples from the P19 (ICU) dataset, with half imputed using linear interpolation as real measured samples and the other half imputed using our model as generated samples. Five ICU-experienced clinicians (3 chief physicians and 2 attending physicians) attempt to distinguish between the two types. As shown in Table \ref{expert}, the experts achieve prediction accuracy of 50.0\%, 48.3\%, 48.3\%, 58.3\%, and 60.0\%, resulting in a kappa score of -0.03. These results are close to random guessing, suggesting that the experts generally struggle to differentiate between the samples. A brief interview further revealed why experts struggled to identify clear patterns to distinguish real from generated samples. One reason is that the complex events in the ICU environment make the data distribution more tolerant. For example, sedation or anesthesia can cause body temperature to fall below the usual range.

\begin{table}[t]
\setlength{\heavyrulewidth}{1.2pt}
\setlength{\tabcolsep}{1mm}
\centering
% \resizebox{0.98\columnwidth}{!}{ 
\begin{tabular}{c ccccc}
\toprule
\multirow{2}{*}{Experts}  & \multicolumn{5}{c}{P19} \\ 
\cmidrule{2-6} 
 & Accuracy & Precision & Recall & F1 score & Specificity\\
\midrule
P1  &50.0  &47.4  &64.3  &54.5 &0.38 \\
P2  &48.3  &44.8  &46.4  &45.6 &0.50 \\
P3  &48.3  &45.5  &53.6  &49.2 &0.44 \\
P4  &58.3  &55.2  &57.1  &56.1 &0.59 \\
P5  &60.0  &57.1  &57.1  &57.1 &0.63 \\
\bottomrule
\end{tabular}
% }
\caption{Performance metrics of five ICU-experienced medical experts.}
\label{expert}
\end{table}

\subsubsection{Ablation study.}
In this section, we present the results of our ablation study in Table \ref{ablation}. The ``default'' one is our standard setup, which includes the sequence encoder, the image encoder, and the joint representation module, along with three self-supervised learning strategies. In the first part of Table \ref{ablation}, we evaluate the performance of individual components: ``image'' signifies that only the image encoder is used for classification, whereas ``sequence'' denotes the use of only the sequence encoder. As a result, we verify that incorporating both sequence and image information significantly improves classification performance, with F1 scores increasing by 2.6\% and 4.5\%. ``sequence-MSE'' denotes the use of MSE loss as the reconstruction loss. In contrast, by replacing it with NT-Xent, we achieved improvements in Accuracy, Precision, Recall, and F1 score by 0.8\%, 0.6\%, 0.2\%, and 0.5\%, respectively. 

The second part of Table \ref{ablation} focuses on our joint representation module. In the ``concatenation'' setting, we simply concatenate sequence and image representations for further downstream classification, and the performance is slightly higher than either ``sequence'' and ``image''. The ``contrastive'' setting shows the improvement from contrastive learning strategy, with 1.1\%, 0.9\%, 1.3\%, and 1.0\% in Accuracy, Precision, Recall, and F1 score. ``Clustering'' strategy also shows positive performance, with 1.2\% in Accuracy, 0.7\% in Precision, 0.9\% in Recall and 0.8\% in F1 score.

\begin{table}[t]
\setlength{\heavyrulewidth}{1.2pt}
\centering
\setlength{\tabcolsep}{0.8mm}
% \resizebox{0.98\columnwidth}{!}{ 
\begin{tabular}{c cccc}
\toprule
\multirow{2}{*}{Methods}  & \multicolumn{4}{c}{PAM} \\ 
\cmidrule{2-5} 
 & Accuracy & Precision & Recall & F1 score \\
\midrule
image  &95.4 {$\pm$ \scriptsize 0.6}  &96.5 {$\pm$ \scriptsize 0.6}   &95.4 {$\pm$ \scriptsize 0.4}  &95.9 {$\pm$ \scriptsize 0.5} \\
sequence &93.3 {$\pm$ \scriptsize 1.1}  & 94.4 {$\pm$ \scriptsize 0.7} &93.6 {$\pm$ \scriptsize 0.6}  & 94.0 {$\pm$ \scriptsize 0.7} \\
sequence-MSE &92.5 {$\pm$ \scriptsize 0.4}  &93.8 {$\pm$ \scriptsize 0.4}  & 93.4 {$\pm$ \scriptsize 0.4} & 93.5 {$\pm$ \scriptsize 0.4} \\
\midrule
concatenation  &95.7 {$\pm$ \scriptsize 0.7}  &96.7 {$\pm$ \scriptsize 0.5}  &96.1 {$\pm$ \scriptsize 0.4}  &96.5 {$\pm$ \scriptsize 0.5}  \\
contrastive  & 96.8 {$\pm$ \scriptsize 0.7} & 97.6 {$\pm$ \scriptsize 0.5} & 97.4 {$\pm$ \scriptsize 0.7} &97.5 {$\pm$ \scriptsize 0.5}  \\
clustering & 96.9 {$\pm$ \scriptsize 0.3} & 97.4 {$\pm$ \scriptsize 0.6} & 97.0 {$\pm$ \scriptsize 0.5}  &97.3 {$\pm$ \scriptsize 0.6}  \\
\midrule
default  &98.3 {$\pm$ \scriptsize 0.3}  &98.7 {$\pm$ \scriptsize 0.6}  & 98.4 {$\pm$ \scriptsize 1.0} & 98.5 {$\pm$ \scriptsize 0.7} \\
\bottomrule
\end{tabular}
% }
\caption{Ablation studies on different strategies.}
\label{ablation}
\end{table}

\section{Conclusion}

In this paper, we propose a joint learning approach of leveraging both sequence and image representations to tackle the classification of irregularly sampled clinical time series. By employing our three self-supervised learning strategies, we are able to effectively learn more generalized joint representations. The effectiveness of our approach is verified on three real-world clinical datasets, where it demonstrates superior performance compared to seven state-of-the-art methods. Additionally, we test our approach under more severe missing rates using leave-sensors-out and leave-samples-out techniques. Our approach consistently achieved strong results, demonstrating its robustness in these scenarios. Our code and data will be made publicly available later. 

\section*{Acknowledgements}
We express our sincere gratitude to all the anonymous reviewers for their valuable guidance and suggestions. We also thank Doctor Weihang Hu, Lin Zhang, and all the colleagues from the Intensive Care Unit at Zhejiang Hospital for their contributions to the expert evaluation and for providing us with valuable clinical advice.

\end{document}